\documentclass[10pt, conference]{IEEEtran}
\usepackage{microtype}
\DisableLigatures{encoding = *, family = * }
\usepackage{mathptmx} 
\usepackage{newtxtext}
\usepackage{cite}
\usepackage{multirow}
\usepackage{booktabs}
\usepackage{rotating}
\usepackage{nicematrix}
\usepackage{amsmath,amssymb,amsfonts}
\usepackage{algorithmic}
\usepackage{graphicx}
\usepackage{tabularx}
\usepackage{textcomp,balance}
\usepackage{xcolor}
\usepackage[ruled]{algorithm2e}
\usepackage{pgfplots}
\usepackage{float}

\def\BibTeX{{\rm B\kern-.05em{\sc i\kern-.025em b}\kern-.08em
    T\kern-.1667em\lower.7ex\hbox{E}\kern-.125emX}}
\usepackage{makecell}
\usepackage{cite}
\usepackage[english]{babel}
\usepackage{amsmath,amssymb,amsfonts}
\usepackage{graphicx}
\usepackage{caption}
\usepackage{lipsum}
\usepackage{cuted}
\usepackage{textcomp}
\usepackage{german}
\usepackage{float}
\usepackage{setspace}
\usepackage{subcaption}
\usepackage{blindtext}
\usepackage{stfloats}
\usepackage{lipsum}
\usepackage{mathtools}
\usepackage{cuted}
\usepackage{booktabs}

\usepackage{amsmath}
\usepackage{mathtools, nccmath}
\newcommand{\RN}[1]{%
  \textup{\uppercase\expandafter{\romannumeral#1}}%
}

\newcommand{\comment}[1]{}

\AtBeginDocument{% 
}   
 \usepackage{booktabs}
 \usepackage{multirow}
\makeatletter
\def\ps@IEEEtitlepagestyle{%
  \def\@oddhead{\hfill\textcolor{red}{\small Accepted for publication in the IEEE Military Communications Conference (IEEE MILCOM 2024)}\hfill}%
  \def\@oddfoot{}%
}
\makeatother
\begin{document}
\addtolength{\topmargin}{+0.01in}
\setlength{\columnsep}{0.25in}
%\addtolength{\footskip}{0.1in}

%\IEEEoverridecommandlockouts\IEEEpubid{\makebox[\columnwidth]{ 978-1-6654-3540-6/22~\copyright~2023 IEEE \hfill} \hspace{\columnsep}\makebox[\columnwidth]{ }}

%\title{ AI/ML-Based Detection of GPS Spoofing Attacks on Military IoT Devices\\

%\title{Optimizing Satellite Retasking in Constellations with Heuristic and Q-Learning Methods}
%\title{Optimizing Satellite Constellation Retasking through Learning Algorithms}

\title{Reinforcement Learning-enabled Satellite Constellation  Reconfiguration and Retasking \\for Mission-Critical Applications}

\author{
\IEEEauthorblockN{
    Hassan El Alami and Danda B. Rawat
}
\IEEEauthorblockA{
    DoD Center of Excellence in AI/ML (CoE-AIML)\\ 
    Howard University, Washington, DC, USA
}
\IEEEauthorblockA{
    %\IEEEauthorrefmark{1}
    \{hassan.elalami, danda.rawat\}@howard.edu}
}

\maketitle

\begin{abstract}
The development of satellite constellation applications is rapidly advancing due to increasing user demands, reduced operational costs, and technological advancements. However, a significant gap in the existing literature concerns reconfiguration and retasking issues within satellite constellations, which is the primary focus of our research. In this work, we critically assess the impact of satellite failures on constellation performance and the associated task requirements. To facilitate this analysis, we introduce a system modeling approach for GPS satellite constellations, enabling an investigation into performance dynamics and task distribution strategies, particularly in scenarios where satellite failures occur during mission-critical operations. Additionally, we introduce reinforcement learning (RL) techniques—specifically Q-learning, Policy Gradient, Deep Q-Network (DQN), and Proximal Policy Optimization (PPO)—for managing satellite constellations, addressing the challenges posed by reconfiguration and retasking following satellite failures. Our results demonstrate that DQN and PPO achieve effective outcomes in terms of average rewards, task completion rates, and response times.
\end{abstract}


\section{Introduction}
%As aerospace technology has advanced rapidly, human endeavors have extended beyond the confines of our atmosphere. 
The emergence of satellite constellations has ushered in a new era in space communications, science, and technology\cite{mensi2023study, rawat2015securing}. Comprising numerous satellites operating in coordination, these constellations have broadened the horizons for military and civilian applications. They enable an unparalleled scale and diversity of operations, from defense to civilian domains such as telecommunications, navigation, and earth observation. As these applications have become widespread, modern society has grown increasingly dependent on satellite infrastructure for tasks like meteorological forecasting, broadband connectivity, and more\cite{lu2023applications}. However, as the number and complexity of satellite missions have grown, so too have the challenges linked to their management. A pressing concern that has recently emerged involves satellites' premature failure or reduced functionality. These malfunctions can be caused by internal issues, such as excessive energy consumption, or external threats, including adversarial attacks. Such complications can render satellites non-operational, leading to gaps in the coverage and services they were designed to provide\cite{yue2023low}. 

According to various studies, despite the rapid advancement in the development of satellite constellations, they face a range of specialized challenges. These challenges include specific orbital arrangements\cite{chen2020service}, high levels of satellite mobility\cite{wang2022efficient}, a vast number of satellites or gateways, resource constraints\cite{huang2020recent}, and issues related to security and reliability\cite{strohmeier2019applicability}. Furthermore, reconfiguration and retasking challenges are of significant concern. Retasking satellites, involving the reallocation or reconfiguration of their functions, is crucial in dynamic space operations. This practice is essential for various reasons, such as accommodating evolving user needs. For example, an Earth observation satellite might shift its focus to respond to a sudden natural disaster\cite{wijata2023taking}. This adaptability also ensures the optimal utilization of onboard resources, such as sensors and bandwidth. Furthermore, if a satellite's primary system fails, retasking can extend the lifespan of its remaining functionalities. The unpredictable nature of space, with challenges like space debris or adverse space weather, may necessitate immediate satellite adjustments\cite{pelton2013space}. As technological advancements unfold, satellites can be retasked by harnessing new techniques or capabilities not originally envisaged. From an economic standpoint, repurposing a satellite often proves more cost-effective than launching a new one\cite{pelton2020conclusion}. In addition, strategic defense imperatives may necessitate the rapid repurposing of satellites to meet emerging reconnaissance or communication needs \cite{weeden2018global}. Components can be reallocated to less demanding roles as they near the end of their lifespan or degrade due to adversarial actions \cite{selva2017distributed}. In multi-satellite constellations, retasking is critical; non-operational satellites can be replaced by repurposing others to ensure mission continuity \cite{brown2018heuristic}.

Recently, several techniques have been introduced to address the challenges of satellite retasking, focusing on reconfiguring satellites either directly within their constellations or through ground stations on Earth. In this context, the authors in\cite{servidia2021autonomous} studied the autonomous reconfiguration of SAR satellite formations using continuous control to tackle the challenge of reconfiguring a satellite cluster during its formation. The study in \cite{jiaxin2021optimal} investigated multi-objective optimization for constellation reconfiguration. It highlights an adaptive multi-objective evolutionary algorithm designed for complex, high-dimensional decision-making contexts. Additionally, the study thoroughly explores system-level scheduling related to constellation reconfiguration, emphasizing its combinatorial strategy model. In\cite{zhang2018self}, a systematic strategy for self-reconfiguration planning tailored for cellular satellites is introduced. An assembling cell is employed to handle or adjust unit cells, ensuring self-reconfiguration capability. The work in \cite{bai2022reconfiguration} delved into optimizing reconfiguration strategies for the formation flight of satellites in elliptical orbits. Existing solutions for satellite constellation reconfiguration have significant limitations, and no comprehensive approach has yet been identified. While many techniques address constellation reconfiguration, there is limited exploration of retasking methods using machine learning (ML). To address this gap, we propose utilizing reinforcement learning (RL) to manage satellites, particularly to handle malfunctions efficiently. Our primary objective is to ensure continuous constellation activity by reallocating tasks from failing satellites to operational ones, thus preventing disruptions. This requires precise task selection during each operational phase to maintain functionality and address issues arising from satellite degradation or sudden failures. Additionally, given the vulnerability of satellites to external threats, rapid retasking is crucial not only for maintaining ongoing operations but also for enhancing system security. The key contributions of our work include:
\begin{itemize}
  \item We propose %a system model and
  optimization problem for GPS satellite constellations, focusing on reconfiguration and retasking, especially in scenarios of satellite failures.
  \item We develop a simulation environment using Python to simulate GPS satellite constellations and perform reconfiguration and retasking in case of satellite failures.
  \item We propose using RL techniques including Q-learning, Policy Gradient methods, DQN, and PPO, for the reconfiguration and retasking of satellite constellations.
  \item We evaluate the performance of the selected RL techniques, assessing their effectiveness based on average reward, task completion rate, and average response time.
\end{itemize}
\noindent The remainder of this paper is organized as follows: Section II describes the system modeling and problem formulation. Section III presents the RL-based algorithms for retasking and reconfiguration of satellite constellations. Section IV provides the results and performance analysis. Finally, Section V concludes the paper.

%\begin{figure*}[htbp]

 %   \centering
 %       \includegraphics[width=0.75\textwidth]{model.jpg}%.65or .77\textewidth
 %   \caption{Space integrated satellite constellation model.}
 %   \label{Fig:model}
%    \vskip-0.45cm
%\end{figure*}

\section{System Modeling and Problem Formulation}
The development of a system model for a constellation of GPS satellites involves several key parameters and functions to manage reconfiguration and retasking in the event of satellite failures. Satellite constellations orbit the Earth in elliptical paths, with their motion described by Keplerian orbital elements. Each satellite \( i \) in the constellation is characterized by a state vector \( S_i \) that includes its position \((x_i, y_i, z_i)\), velocity \((v_{x_i}, v_{y_i}, v_{z_i})\), operational status \(o_i\), energy level \(e_i\), and task load \(T_i\). The operational status \(o_i\) is binary, indicating whether the satellite is operational '1' or failed '0'. The failure probability \( p_i \) can be modeled as a function of time and environmental factors, impacting the satellite's operational status. Communication systems in satellite constellations are crucial for coordinating tasks, transmitting data, and ensuring the overall functionality of the satellite constellation. Each satellite must maintain communication with both other satellites in the constellation and ground stations. Key communication parameters include bandwidth \( B_{ij} \) between satellites \( i \) and \( j \), operating frequency \( f_{ij} \), and modulation scheme \( m_{ij} \). Task distribution, a crucial component of the model, involves redistributing the total task load \( T_{\text{total}} \) among the remaining operational satellites \( \mathcal{O} \). When a satellite \( j \) fails, its tasks \( T_j\) are redistributed among the remaining operational satellites. The task redistribution is given by \( T_i' = T_i + \frac{T_j}{|\mathcal{O}|} \) for all \( i \) in \( \mathcal{O} \). Reconfiguration and retasking are performed when a satellite fails, involving recalculating positions and adjusting tasks. The new positions and velocities of the satellites are updated using equations of motion over a time interval \( \Delta t \). For position updates: \( x_i(t + \Delta t) = x_i(t) + v_{x_i}(t) \Delta t \), and for velocity updates considering possible thrust adjustments: \( v_{x_i}(t + \Delta t) = v_{x_i}(t) + a_{x_i}(t) \Delta t \).\\

Performance metrics are crucial for evaluating the satellite constellation. The Task Completion Rate (TCR) is a percentage-based measure that assesses how effectively a satellite constellation completes its assigned tasks. A higher TCR signifies efficient task execution, while a lower rate may indicate inefficiencies or issues within the satellite constellation. It is calculated as:

%\noindent{\text{Task Completion Rate (TCR)}}:
\begin{align}
\text{TCR} &= \frac{\text{Initial tasks} - \text{Remaining tasks}}{\text{Initial tasks}} \times 100 \label{eq:TCR}
\end{align}

The Average Response Time (ART) measures the system's responsiveness during satellite failures. ART quantifies the average time required to detect a satellite failure, retask, and reconfigure the remaining operational satellites to handle the tasks previously managed by the failed satellite. This includes the entire process from detecting the satellite failure to successfully redistributing the tasks among the remaining satellites. A lower ART indicates a more responsive and adaptable system, capable of quickly adjusting to failures and maintaining operational efficiency. Conversely, a higher ART suggests potential delays and inefficiencies in the retasking and reconfiguration processes. ART is defined as (\ref{eq:ART}).

%\noindent{\text{Average Response Time (ART)}}:
\begin{align}
\text{ART} &= \frac{\sum_{i=1}^{N} t_i}{N} \label{eq:ART}
\end{align}
where \( t_i \) is the response time for the \( i \)-th satellite failure (the time taken to detect the failure and complete the reconfiguration and retasking) and \( N \) is the total number of satellite failures.\\

The optimization problems in this context focus on maximizing the TCR and minimizing the ART. Specifically, the optimization problems can be formulated as follows, with their corresponding constraints:\\

\noindent{\text{Maximizing TCR}}:
\begin{equation}
\begin{cases}
   \max \text{TCR} = \frac{\text{Initial tasks} - \text{Remaining tasks}}{\text{Initial tasks}} \times 100\% \\
   \text{s.t.,}~
   c_i \leq C_i \quad \forall i \in \mathcal{O}
\end{cases}
\label{eq:max_TCR}
\end{equation}
where \( C_i \) is the capacity of operational satellite \( i \).\\

\noindent{\text{Minimizing ART}}:
\begin{equation}
\begin{cases}
   \min \text{ART} = \frac{\sum_{i=1}^{N} t_i}{N} \\
   \text{s.t.,}~
   t_i \leq t_{\text{max}} \quad \forall i
\end{cases}
\label{eq:min_ART}
\end{equation}
where \( t_{\text{max}} \) is defined as the maximum allowable time taken to detect the failure and complete the reconfiguration and retasking for each satellite failure.\\

These optimization problems aim to enhance the overall performance and reliability of the satellite constellation in mission-critical applications by ensuring efficient task redistribution and minimizing response times in the event of satellite failures.
%\vskip-0.23cm
\begin{algorithm}[!htpb]
\small
\caption{Satellite Constellation Environment}
\label{alg:constellation}
\textbf{Initialize:}\\
$N$: Number of satellites\\
$R$: Number of rounds\\
$action\_space$: Action space\\
$observation\_space$: Observation space\\
$B$: Bandwidth matrix\\
$F$: Frequency matrix\\
$p$: Failure probability\\
$episode\_counter$: Episode counter\\

\textbf{Function} \textit{initialize\_satellites()} \\
\begin{itemize}
    \item \textbf{for each satellite $i$ in $N$}:
    \begin{itemize}
        \item Initialize properties: task load, maximum capacity, energy level, bandwidth capacity, frequency modulation, status, reliability.
        \item Calculate initial orbital positions: inclination, angles, radius, x, y, z coordinates.
    \end{itemize}
\end{itemize}

\textbf{Function} \textit{reset()} \\
\begin{itemize}
    \item Set current round to 0.
    \item Reset properties of non-failed satellites.
    \item Identify and redistribute tasks from failed satellites.
    \item Set initial tasks and metrics.
    \item \textbf{if} $episode\_counter \geq 100$:
    \begin{itemize}
        \item Increase $p$ based on episode count.
    \end{itemize}
    \item Increment $episode\_counter$.
    \item \textbf{return} initial state.
\end{itemize}

\textbf{Function} \textit{calculate\_distance($sat_i$, $sat_j$)} \\
\begin{itemize}
    \item Calculate Euclidean distance between satellites $sat_i$ and $sat_j$.
\end{itemize}

\textbf{Function} \textit{step(action)} \\
\begin{itemize}
    \item Parse $action$ into $from\_sat$ and $to\_sat$.
    \item \textbf{if} $from\_sat$ or $to\_sat$ not operational or $from\_sat$ equals $to\_sat$:
    \begin{itemize}
        \item \textbf{return} current state, penalty reward, and $done$ flag.
    \end{itemize}
    \item Calculate distance, delay, bandwidth, and frequency modulation.
    \item Compute effective task transfer and update satellite states.
    \item Apply penalties and rewards.
    \item \textbf{if} a satellite fails:
    \begin{itemize}
        \item Redistribute tasks among operational satellites.
        \item Reconfigure the constellation.
    \end{itemize}
    \item Update parameters.
    \item Increment current round.
    \item \textbf{return} new state, reward, and $done$ flag.
\end{itemize}
\end{algorithm}
%\vskip-0.3cm
\section{RL-Based Algorithms for Retasking and Reconfiguration in Satellite Constellation}
This section describes the constellation environment and the RL models used to develop learning strategies for satellite reconfiguration and retasking within satellite constellations. The Algorithm \ref{alg:constellation} outlines the operation of a satellite constellation environment, typically used for satellite reconfiguration and retasking in mission-critical applications. It begins by initializing parameters such as the number of satellites $N$, the number of rounds $R$, the action space, the observation space, the bandwidth matrix $B$, the frequency matrix $F$, the failure probability $p$, and an episode counter. In mission-critical applications, the probability $p$ of satellite failures is assumed to be higher due to the increased operational demands and the harsh conditions that these satellites often encounter. The \textit{initialize\_satellites()} function sets up each satellite's properties and calculates their initial orbital positions. These properties include task load, maximum capacity, energy level, bandwidth capacity, frequency modulation, status, and reliability. The \textit{reset()} function resets the environment for a new episode. It sets the current round to 0, restores properties of non-failed satellites, redistributes tasks from failed satellites and adjusts parameters based on the episode count. For example, if the episode counter is greater than or equal to 100, the failure probability is increased. The \textit{calculate\_distance()} function computes the Euclidean distance between two satellites based on their positions. This distance calculation is essential for determining various aspects of satellite communication and task execution. The \textit{step(action)} function is the core of the algorithm, facilitating interaction between the environment and an agent. It parses the action into source and destination satellites, checks for validity, calculates parameters such as distance, delay, bandwidth, and frequency modulation, and executes the action. The function also handles penalties and rewards and manages responses to satellite failures. If a satellite fails, tasks are redistributed among operational satellites, and the constellation is reconfigured to ensure continued operation. Finally, the function updates metrics and returns the new state, reward, and a flag indicating whether the episode is done.
\begin{table*}[!b]
\centering
\caption{Tuning parameters for simulation for selected RL techniques}
\label{table:tuning_parameters}
\begin{tabularx}{\textwidth}{@{}l*{4}{>{\centering\arraybackslash}X}@{}}
\toprule
\textbf{Parameter} & \textbf{Q-Learning} & \textbf{Policy Gradient} & \textbf{DQN} & \textbf{PPO} \\ 
\midrule
Learning Rate & 0.1 & 0.001 & 0.0001 & 0.0003 \\
Discount Factor ($\gamma$) & 0.99 & 0.99 & 0.99 & 0.99 \\
Exploration Rate ($\epsilon$) & 0.1 & - & 0.1 (decay) & - \\
Batch Size & - & 32 & 64 & 64 \\
Replay Buffer Size & - & - & 10000 & - \\
Target Network Update Frequency & - & - & 1000 steps & - \\
Entropy Coefficient & - & 0.01 & - & 0.01 \\
Gradient Clipping & - & 0.5 & - & 0.5 \\
Policy Update Frequency & - & 1 & - & 4 \\
\bottomrule
\end{tabularx}
\end{table*}
%\vskip-0.7cm
\begin{algorithm}[!b]
\small
\caption{Interacting with a Satellite Constellation}
\label{alg:rl_interaction}
\textbf{Initialize:}\\
Satellites Environment: \textbf{Algorithm \ref{alg:constellation}}\\
$episode\_counter$: Episode counter\\
$RL$: Selected RL models including Q-learning, Policy Gradient, DQN, PPO\\

\textbf{Define:}\\
$TCR$: Task Completion Rate\\
$RT$: Response Time\\

\textbf{Function} \textit{initialize\_environment()} \\
\begin{itemize}
    \item Initialize the satellite environment with parameters.
\end{itemize}

\textbf{Function} \textit{initialize\_agent()} \\
\begin{itemize}
    \item Initialize RL agents:
    \begin{itemize}
        \item Initialize Q-table for Q-learning.
        \item Initialize Policy Network for Policy Gradient.
        \item initialize NN that approximates the Q-values in DQN.
        \item Initialize Policy Network for PPO.
    \end{itemize}
\end{itemize}

\textbf{Function} \textit{train\_agents()} \\
\begin{itemize}
    \item Repeat for a fixed number of episodes:
    \begin{itemize}
        \item Reset the environment to its initial state.
        \item Reset the episode-specific variables and counters.
        \item Repeat for each round:
        \begin{itemize}
            \item Observe the current state of the environment.
            \item Choose actions using the RL model (Q-learning, Policy Gradient, DQN, or PPO) based on the observed state.
            \item Execute the chosen actions in the environment.
            \item Receive rewards and observe the next state.
            \item Update the RL model's parameters based on the observed transition (state, action, reward, next state).
            \item \textbf{Handle satellite failure:}
            \begin{itemize}
                \item Detect any satellite failures.
                \item Perform retasking and reconfiguration of the remaining operational satellites.
                \item Update the environment to reflect these changes.
            \end{itemize}
        \end{itemize}
    \end{itemize}
\end{itemize}

\textbf{Function} \textit{optimize\_performance()} \\
\begin{itemize}
    \item Analyze the performance of RL agents based on average rewards, task completion rate, and response time.
    \item Adjust hyperparameters of RL algorithms iteratively to improve performance.
\end{itemize}
\end{algorithm}
In Algorithm \ref{alg:rl_interaction}, the selected RL models, including Q-learning, Policy Gradient, DQN, and PPO, are integrated into a satellite constellation environment to enhance the average reward, TCR, and ART. The algorithm begins by initializing key parameters, including the number of satellites, rounds, action, observation spaces, and the episode counter. For each RL model, the essential components are set up: the Q-table for Q-learning, the policy network for the Policy Gradient method, the neural network (NN) for the DQN, and the policy network for PPO. The process involves several key functions: initializing the environment and agents, training the agents through interaction with the environment, and optimizing their performance based on specific metrics. During the training phase, the agents operate over a series of episodes. In each episode, the environment is reset to its initial state, and agents' episode-specific variables and counters are also reset. The agents then engage in multiple rounds within each episode, where they observe the current state, choose actions based on their respective RL models, execute those actions, receive rewards, and update their parameters based on the observed transitions (state, action, reward, next state). Additionally, the algorithm handles satellite failures by detecting any such failures, performing retasking and reconfiguration of the remaining operational satellites, and updating the environment accordingly. To optimize performance, the algorithm evaluates the agents' effectiveness using metrics such as average rewards, TCR, and ART. Based on these evaluations, hyperparameters of RL algorithms can be iteratively adjusted to improve performance. This systematic framework ensures that RL are effectively trained to optimize key performance indicators in the satellite environment, enhancing both task completion rates and response times. Overall, the algorithm provides a comprehensive approach to incorporating RL into the satellite environment, focusing on improving operational efficiency through systematic training and performance optimization.

\section{Performance Evaluation}
In this section, we perform a comprehensive assessment comparing the performance of load balancing, Q-learning, Policy Gradient, DQN, and PPO in satellite reconfiguration and retasking scenarios. The evaluation focuses on average rewards, task completion rate, and average response time.
\subsection{Simulation Settings}
In our study, we developed a Python-based simulation of a 24-satellite GPS constellation operating on three frequency bands (L1, L2, L5). The tuning parameters used for the selected RL models to optimize their performance in the simulation are presented in Table~\ref{table:tuning_parameters}. The selected techniques are evaluated based on task completion rate (TCR) and average response time (ART), defined by (\ref{eq:TCR}) and (\ref{eq:ART}), respectively.
\subsection{Empirical Results}

\begin{figure}[!t]
    \centering
   % \vskip-.65cm
\includegraphics[width=\columnwidth]{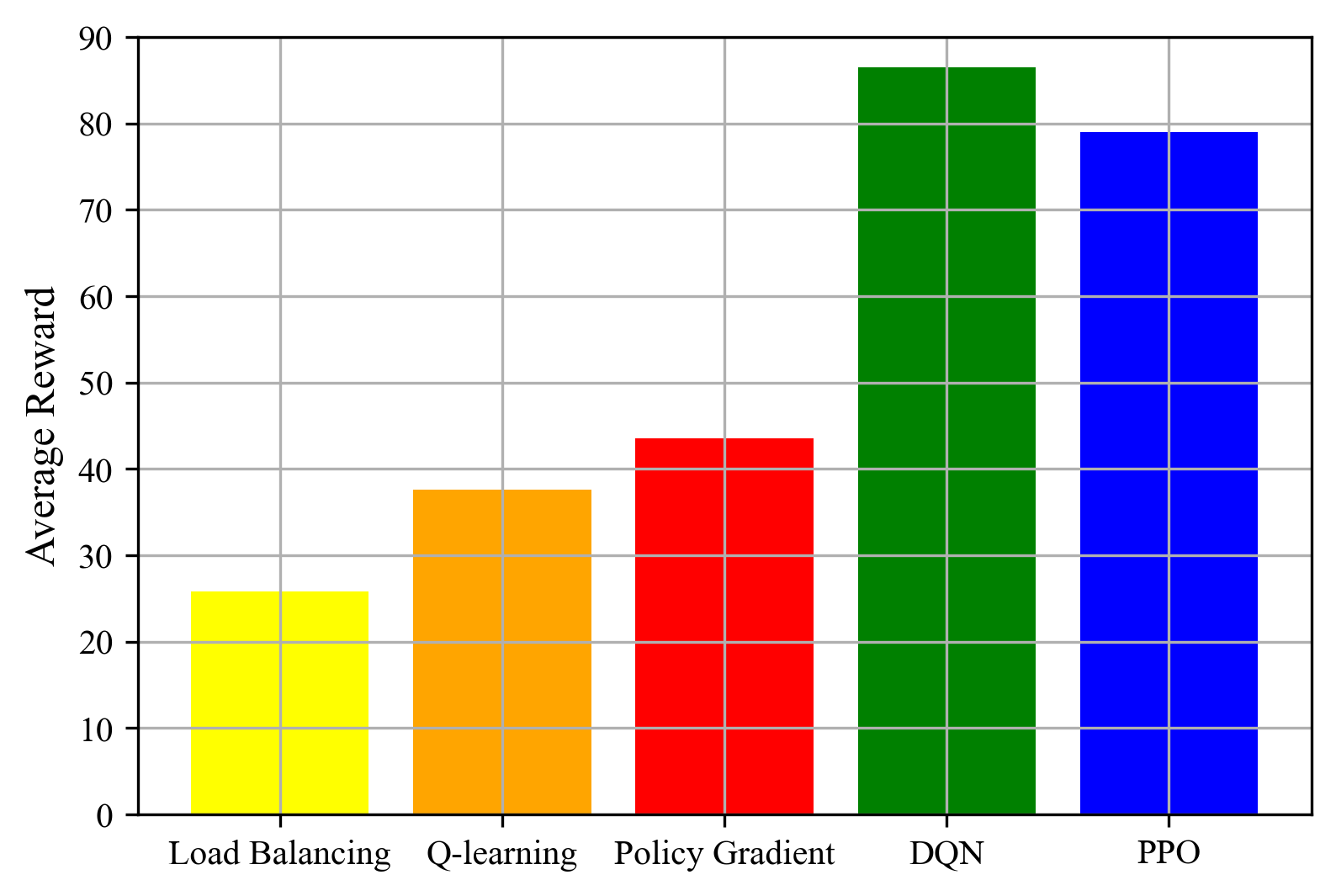}
    \caption{Average rewards of selected techniques.}
    \label{task:rw}
       %\vskip-.5cm
\end{figure}

As presented in Fig.~\ref{task:rw}, the comparative analysis of average rewards for retasking in satellite constellation reveals significant differences in the effectiveness of selected techniques. Load Balancing, with an average reward of 25.75, demonstrates its limitations in dynamically optimizing satellite task distribution. Although it ensures equitable workload distribution, it fails to adapt to the complex and variable environment needed for maximizing overall rewards. RL techniques, such as Q-learning and Policy Gradient, show incremental improvements, with average rewards of 37.6 and 43.49, respectively. Q-learning's iterative value updating enhances action selection over time, and Policy Gradient's direct policy optimization is advantageous for continuous action spaces. However, these methods are outperformed by more advanced techniques, indicating potential for further enhancement. DQN, with a substantial average reward of 86.5, illustrates the transformative potential of deep learning within RL frameworks. DQN's use of neural networks to approximate Q-values facilitates effective management of high-dimensional state and action spaces, resulting in superior average rewards. This significant performance improvement highlights DQN's capability to learn and generalize complex patterns, making it exceptionally suitable for the dynamic demands of satellite retasking. PPO also demonstrates robust performance with an average reward of 79, emphasizing its stability and efficiency. PPO maintains training stability through clipped probability ratios, preventing disruptive updates that could destabilize the learning process. This stability, combined with high performance, makes PPO a compelling choice in scenarios where training robustness is critical. The results highlight the superior performance of DQN and PPO compared to the load balancing approach and basic RL models such as Q-learning and Policy Gradient, emphasizing the critical importance of advanced RL techniques in optimizing satellite operations. DQN emerges as the most effective method for maximizing average rewards, while PPO offers a balanced approach with stability and high rewards. These findings advocate for the integration of advanced RL techniques in practical satellite retasking applications, promising substantial enhancements in operational efficiency and effectiveness.\\

\begin{figure}[!t]

    \centering
       % \vskip-.25cm
\includegraphics[width=\columnwidth]{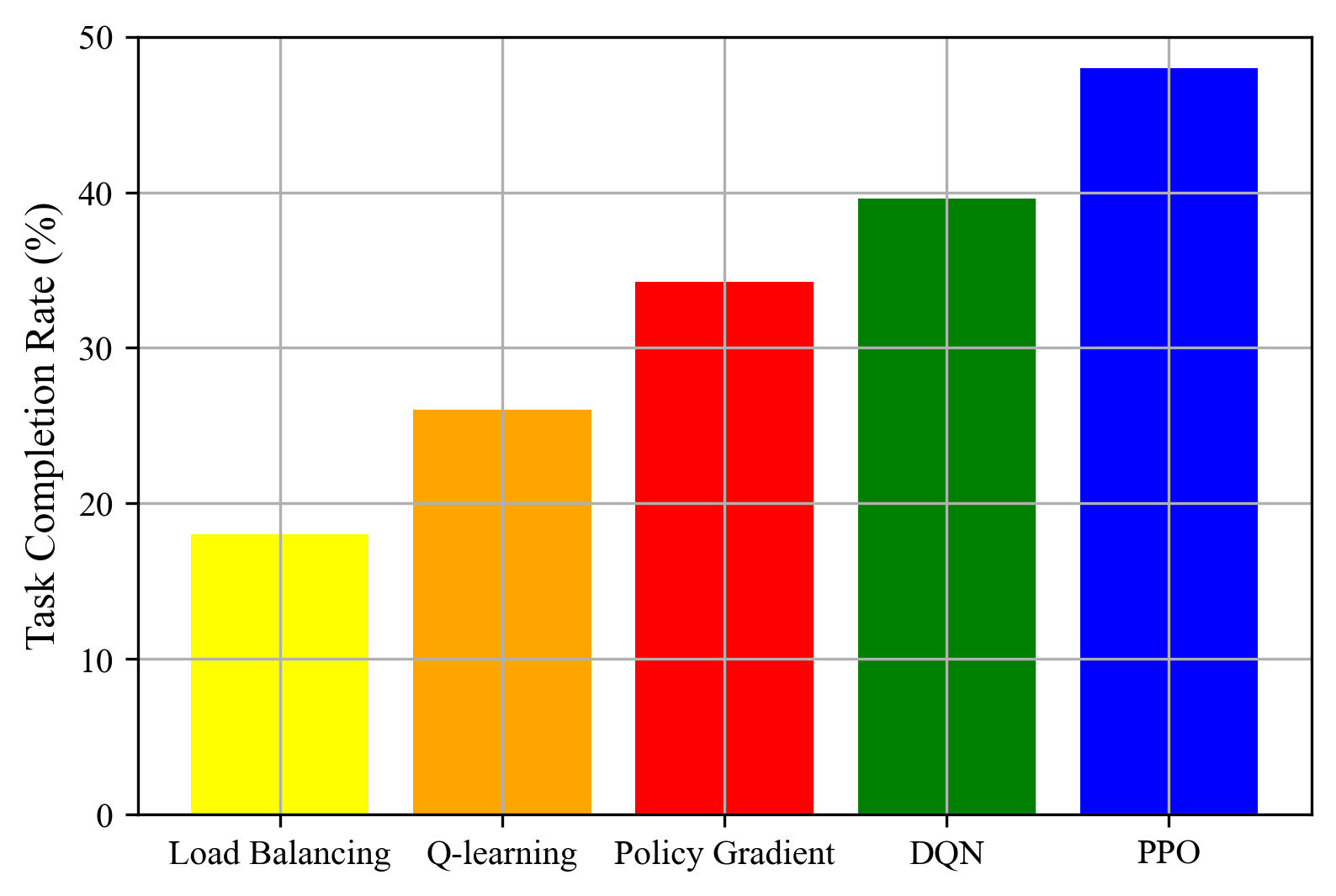}
    \caption{Completion tasks rate of selected techniques.}
    \label{task:rl}
        %\vskip-.45cm
\end{figure}

As illustrated in Fig.~\ref{task:rl}, the Load Balancing method, with a task completion rate of 18\%, is the simplest method and exhibits the lowest performance due to its static approach to task distribution, which fails to adapt to changing conditions or satellite network failures. Q-learning, achieving a task completion rate of 26\%, is an early RL algorithm that demonstrates some improvement through learning optimal actions via trial and error; however, its effectiveness is limited by the absence of advanced techniques. The Policy Gradient method, with a task completion rate of 34.25\%, enhances performance by directly optimizing the policy, thereby facilitating more nuanced decision-making that adapts better to environmental changes. DQN, incorporating deep neural networks, significantly improves task completion to 39.60\% by learning more complex representations of the state-action space, enabling superior decision-making in dynamic scenarios. PPO achieves the highest task completion rate at 48\%, significantly surpassing the other methods. This superior performance is attributed to PPO's sophisticated optimization techniques, such as clipped probability ratios and trust region updates, which ensure stable and efficient learning by preventing large, destabilizing updates. PPO’s capability to balance exploration and exploitation efficiently allows it to adapt rapidly to the dynamic and unpredictable nature of the satellite environment, particularly in scenarios involving satellite failures and task redistributions. This robustness renders PPO exceptionally effective at maintaining high performance in complex environments, showcasing its superior capability in managing and completing tasks under varying operational conditions. Consequently, PPO's advanced techniques and adaptive learning capabilities establish it as the most effective RL method for optimizing task completion rates in the context of satellite management.\\

\begin{table}[!h]
\centering
\caption{Performance comparison of average response time of selected techniques}
\label{table:time}
\begin{tabularx}{\columnwidth}{@{}l>{\centering\arraybackslash}X@{}}
\toprule
\textbf{Model} & \text{Average Response Time (s)} \\ 
\midrule
Load Balancing & 1  \\
Q-Learning & 0.99  \\
Policy Gradient & 0.985 \\
DQN & \textbf{0.907} \\
PPO & 0.923 \\
\bottomrule
\end{tabularx}
\end{table}
    %\vskip-.75cm
The performance comparison of average response times for the proposed techniques in reconfiguration and retasking of satellite constellation, as presented in Table \ref{table:time}, shows distinct efficiency differences. Load Balancing, with an average response time of 1 second, is the least efficient, indicating significant delays. Q-Learning improves slightly with 0.99 seconds, suggesting marginal enhancements due to adaptive learning. Policy Gradient methods further reduce response time to 0.985 seconds, benefiting from direct policy optimization. The most notable improvements are seen with DQN and PPO. DQN achieves the lowest response time of 0.907 seconds, highlighting its superior capability to handle complex state-action spaces via neural networks. PPO, with an average response time of 0.923 seconds, emphasizes stability and training efficiency. These results underscore the importance of advanced reinforcement learning techniques in optimizing satellite operations. DQN and PPO, with their quick response, stability, and efficiency, represent significant advancements over the load balancing method and conventional RL techniques. Their integration into satellite retasking applications promises substantial improvements in operational efficiency and responsiveness. DQN and PPO are particularly effective in retasking and reconfiguring satellite constellation, especially during failures, due to their advanced algorithms. DQN’s use of deep neural networks, experience replay, and target networks ensures robust policy development and adaptation to dynamic conditions. PPO's balance between exploration and exploitation, suitability for continuous action spaces, and sample efficiency enhance performance. 

\section{Conclusion and Future Work}
This paper addresses the challenges of reconfiguration and retasking in the event of satellite failures within a GPS satellite constellation for mission-critical applications. We propose optimization problems specific to the GPS satellite constellation and introduce load balancing along with RL models, including Q-learning, Policy Gradient, DQN, and PPO. Subsequently, we conduct a comprehensive performance assessment of the selected techniques, considering average rewards, task completion rates, and average response times as evaluation criteria. The results show that RL models, particularly DQN and PPO, outperform other techniques due to their dynamic adaptability. Both techniques optimize resource utilization and ensure reliable decision-making during satellite failures, making them highly suitable for maintaining constellation operations. Their implementation in satellite management systems is anticipated to significantly enhance operational resilience and efficiency. Future research will focus on refining the proposed system model and simulation environment to closely replicate real-world satellite constellations. Additionally, we plan to explore the applicability of advanced AI algorithms, including large language models (LLMs), to enhance the effectiveness of RL algorithms in satellite constellation management.

\section*{Acknowledgment}
This work was supported in part by the DoD Center of Excellence in AI and Machine Learning (CoE-AIML) at Howard University under Contract W911NF-20-2-0277 with the U.S. Army Research Laboratory, VMware Inc. Research Gift Funds and Microsoft Corp. Research Gift Funds. % as well as by the NSF grants CNS 1650831. 

%\bibliographystyle{IEEEtran}
%\bibliography{main}
%\end{document}
\bibliographystyle{IEEEtran}
%\bibliography{main}  % Comment this out

  % Include this instead
\end{document}